\documentclass[10pt,twocolumn,letterpaper]{article}

\usepackage{FT}              %

\usepackage{multirow}
\usepackage{float}
\usepackage{amssymb} % For checkmark symbols
\usepackage{xcolor}  % For coloring symbols
\usepackage{utfsym}

\definecolor{cblue}{rgb}{0.21,0.49,0.74}
\usepackage[pagebackref,breaklinks,colorlinks,allcolors=cblue]{hyperref}
\usepackage{booktabs}
\usepackage{multirow}
\usepackage{multicol}
\usepackage{color, colortbl}
\usepackage{pifont}%
\usepackage{array}
\usepackage[misc]{ifsym} %

\title{FTMoMamba: Motion Generation with Frequency and Text State Space Models}

\author{Chengjian Li$^{1}$ \quad
Xiangbo Shu$^{1~\textrm{\Letter}}$ \quad
Qiongjie Cui$^{1}$ \quad
Yazhou Yao$^{1}$ \quad
Jinhui Tang$^{1~\textrm{\Letter}}$
\vspace{0.3em} \\
\textsuperscript{1} School of Computer Science and Engineering, Nanjing University of Science and Technology 
}
\begin{document}
\maketitle

\let\thefootnote\relax\footnotetext{$^{~\textrm{\Letter}}$ Corresponding author: Xiangbo Shu, Jinhui Tang (\url{shuxb@njust.edu.cn,jinhuitang@njust.edu.cn}).}

\begin{abstract}
   Diffusion models achieve impressive performance in human motion generation. However, current approaches typically ignore the significance of frequency-domain information in capturing fine-grained motions within the latent space (e.g., low frequencies correlate with static poses, and high frequencies align with fine-grained motions). Additionally, there is a semantic discrepancy between text and motion, leading to inconsistency between the generated motions and the text descriptions. In this work, we propose a novel diffusion-based {\textbf{FTMoMamba}} framework equipped with a Frequency State Space Model (FreqSSM) and a Text State Space Model (TextSSM). Specifically, to learn fine-grained representation, {\textbf{FreqSSM}} decomposes sequences into low-frequency and high-frequency components, guiding the generation of static pose (e.g., sits, lay) and fine-grained motions (e.g., transition, stumble), respectively. To ensure the consistency between text and motion, {\textbf{TextSSM}} encodes text features at the sentence level, aligning textual semantics with sequential features. Extensive experiments show that FTMoMamba achieves superior performance on the text-to-motion generation task, especially gaining the lowest FID of 0.181 (rather lower than 0.421 of MLD) on the HumanML3D dataset.
\end{abstract}
\section{Introduction}
% \label{sec:intro}
\textcolor{black}{Human motion generation is the process of creating realistic motion based on given conditions \cite{zhang2025motion}. This task aims to synthesize motions that conform to physical laws and logical motions, e.g., walking, running, jumping, etc. It is widely applied in animation, VR/AR, game development, and human-computer interaction \cite{dai2023slowfast,chen2023executing}. } In the early stage, researchers explore generation methods by leveraging autoencoders \cite{guo2022generating,tevet2022motionclip,zhong2023attt2m}, generative adversarial networks (GANs) \cite{ahn2018text2action,barsoum2018hp}, autoregressive models \cite{jiang2023motiongpt,gong2023tm2d}, etc. They generally face challenges such as information loss, training difficulties, and error accumulation in long-sequence motion generation. Recently, inspired by the Diffusion model that has shown remarkable performance in image and video generation tasks, researchers \cite{chen2023executing, tevet2023human} turn to design various Diffusion-based methods for motion generation.\par

However, most Diffusion-based methods~\cite{zhang2024generative,song2023finestyle} fail to simultaneously capture static poses and fine-grained motions of the motion generation process in the common semantic space. Take the qualitative results in Figure \ref{fig:1}(a) as an example, when executing the first instruction \textit{``A person walks forward, sits"}, a Diffusion-based baseline method (\textit{i.e.}, MLD \cite{chen2023executing}) fails to effectively capture the static pose (\textit{e.g}., sit). For the second instruction \textit{``rise-walk-lay"}, it struggles to generate fine-grained motion (\textit{e.g.}, transition). Here, the transition is the \textit{``rise"} to \textit{``walk"}.  
Inspired by \cite{chang2010frequency} that uses high-frequency information to recognize dynamic actions and \cite{kothandaraman2022differentiable} that leverages low-frequency information to identify non-significant regions, we consider introducing the frequency domain for separately capturing static poses and fine-grained motions in the denoising process of the Diffusion model.

Moreover, most Diffusion-based methods \cite{chen2023executing, zhang2025motion} cannot ensure the consistency of the text-motion semantic information in the spatiotemporal space well due to the intrinsic difference between text descriptions and motion sequences. 
As shown in Figure \ref{fig:1}(a), for the third instruction \textit{``Walking forward and steps over an object, and then continue walking"}, the Diffusion-based baseline method obviously lacks an understanding of \textit{``Object"}, leading to the omission of the \textit{``step"} motion. Here, the above method directly concatenates text features and motion features. Such simple concatenation brings in the text-motion semantic inconsistency when the motions become more complex. Therefore, we consider optimizing the design of the text-conditioned guidance for accurately aligning the texts and motions in the denoising process of the Diffusion model.

In the denoising process of the Diffusion model, accurately predicting noise through the UNet-like denoising architecture becomes mainstream for the quality of motion generation~\cite{huang2024stablemofusion,chen2023executing}. Currently, there are several module options for the UNet-like denoising architecture, including convolution, Transformer, and Mamba~\cite{gu2023mamba}. Among them, Mamba is more computationally efficient compared to Transformer \cite{vaswani2017attention}, while maintaining the same long-range modeling capabilities. Therefore, we consider using Mamba as the foundational module to bring in frequency information and text-conditioned guidance. One straightforward way is to regard the frequency information or text-conditioned guidance as residual connections of the Mamba module. However, such a way may bring in the interfering information to the model, which negatively impacts performance improvement. Some researchers \cite{wan2404sigma,chen1993frequency} have stated that the core components of SSM in Mamba, namely the \textbf{\textit{A}} and \textbf{\textit{C}} matrices, serve the roles of capturing frequency domain pole and decoding hidden state information, respectively. Therefore, a more graceful method would be to incorporate low- and high-frequency information into the \textbf{\textit{A}} matrix and text features into the \textbf{\textit{C}} matrix. \par

To this end, we introduce a new FTMamba, a combination of FreqMamba and TextMamba, which includes a new Frequency State Space Model (FreqSSM) and a new Text State Space Model (TextSSM) in FreqMamba and TextMamba, respectively. Specifically, FreqSSM integrates low-frequency and high-frequency information into the matrix \textbf{\textit{A}}, capturing both local and global motion variations to guide the model in generating static poses and fine-grained motions. TextSSM aligns text features and motion features in the sentence-level matrix $\textbf{\textit{C}}_{\text{s}}$ to ensure consistency between text and motion in the latent space. More technical details are introduced in Section~\ref{sec3.2}. Building upon the FTMamba, we present a novel FTMoMamba framework, which overcomes the challenges of fine-grained motion characterization and text-motion inconsistency. As shown in Figure \ref{fig:1}(b), FTMoMamba achieves better performance, especially when the FID value is significantly reduced to 0.18, while preserving the acceptable Flops and Parames.\par
{\color{black}
Our main contributions are threefold.  {\bf (1) Focus on fine-grained motion generation.} We propose a novel diffusion-based FTMoMamba framework that explores extra frequency-domain information and text-semantic information via FreqSSM and TextSSM to well capture fine-grained motions and ensure the text-motion consistency in the latent space. {\bf (2) Bring frequency-domain information in SSM.} We present a Frequency State Space Model (FreqSSM) that obtains low- and high-frequency information to capture the static poses and fine-grained motions, respectively. {\bf (3) Explore text-motion consistency in SSM.} We propose a Text State Space Model (TextSSM) that extracts the sentence-level features to guide the text to motion alignment, guaranteeing text-motion consistency.
}
% , which unitizes the frequency-domain information and text descriptions to ensure the gerneration of fine-grained motion and text-motion consistecny
\begin{figure*}
    \centering
    \includegraphics[width=1\linewidth]{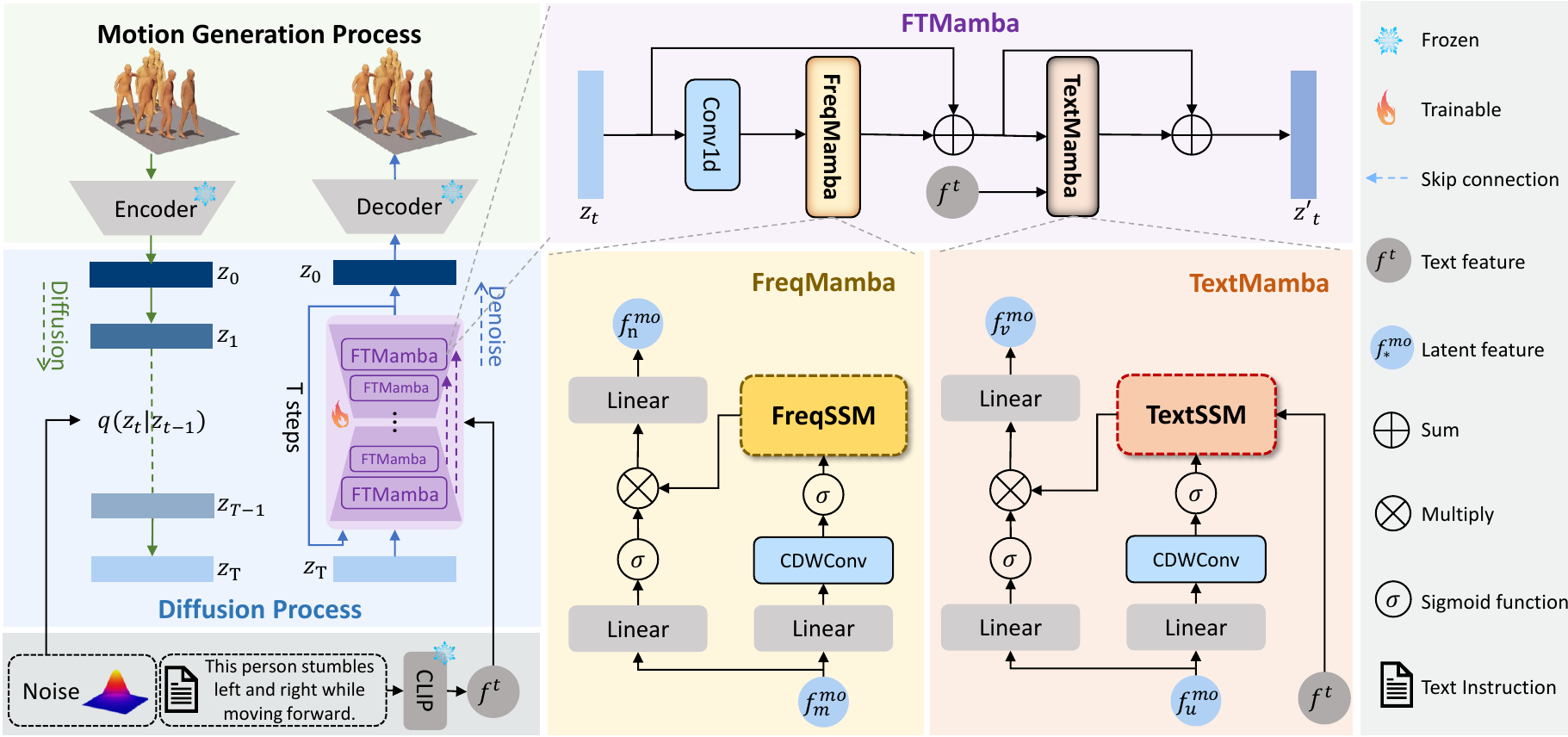}
    \caption{Overview of FTMoMamba. FTMoMamba is built upon the Diffusion model within FTMamba modules, which explores the frequency-domain information to guide motion generation, as well as text-semantic information to ensure text-motion consistency in the latent space. Specifically, the diffusion model compresses and decompresses the raw motion sequence, reducing the interference of redundant information in motion generation. FTMamba, as the core of the denoising module, consists of FreqMamba and TextMamba. The former decomposes motion sequences into low- and high-frequency components to guide the generation of static and fine-grained motions, respectively. The latter aligns textual semantics with sequential features to ensure text-motion consistency.}
    
    \label{fig:2}
    % \vspace{-0.5cm}
\end{figure*}
\section{Related Works}
\noindent{\bf Human Motion Generation.}
Generating human motion is an essential application of computer vision that is widely used in a variety of fields such as 3D modeling and robot manipulation. Recently, the main method for generating human motion has been called the Text-to-Motion task, which processes both language and motion by learning a shared latent space. In this task, autoencoder-based methods \cite{tevet2022motionclip,petrovich2022temos,guo2022generating} robustly represent data by compressing high-dimensional information into a latent space. Autoregressive-based models \cite{jiang2023motiongpt,gong2023tm2d,zhong2023attt2m} generate motions step by step, predicting the next motion based on the previously generated one. Diffusion model-based on latent space \cite{zhang2024motiondiffuse,tevet2023human,yuan2023physdiff,shafir2023human,chen2023executing} generates high-quality motion through a gradual denoising process. Although the above methods have achieved some success, they still face challenges such as high computational cost, difficulty in fine-grained characterization, and text-motion inconsistency. Our work, based on the state space model (SSM), leverages frequency domain information and text-motion alignment to achieve fine-grained characterization and text-motion consistent generation.\par
\noindent{\bf Frequency Domain State Space Model.}
Frequency domain state space models \cite{xu2024demmamba,li2024fouriermamba,huang2024irsrmamba} enhance the model’s image processing capabilities by decomposing images into low- and high-frequency information and utilizing information injection and scanning strategies. Despite these methods being effective, the lack of in-depth analysis of the SSM leads to the introduction of redundant information by simply adding branches and directly scanning the model. Our work analyzes the SSM and finds that the state transition matrix \textbf{\textit{A}} is related to the frequency domain poles but lacks the ability to perceive local and global trends. Therefore, we propose a frequency state space model (FreqSSM), which integrates low-frequency and high-frequency information into the matrix \textbf{\textit{A}}, introducing both local and global trend variations to guide the model in generating static poses and fine-grained motions.
\noindent{\bf Cross-Modal State Space Model.}
The cross-modal state space model \cite{liu2024robomamba,dong2024fusion}  extracts different modality features through multiple branches of Mamba and performs modality feature fusion using cross-attention, concatenation, or addition. Although these methods improve model performance, they overlook the correlation between the matrix \textbf{\textit{C}} in the SSM and cross-modal information fusion, leading to additional computational overhead. We propose a TextSSM that centers around the output matrix \textbf{\textit{C}}, achieving text-motion information fusion with negligible computational cost, ensuring text-motion consistency.\par
\section{Methods}
\label{sec:formatting}
\subsection{Preliminaries}
\textbf{State Space Model (SSM) }is a mathematical framework used to describe dynamic systems, with models like the Structured State Space for Sequence (S4) and Mamba excelling in handling long sequences. These models use a hidden state $\textbf{\textit{h}}(t) \in \mathbb{R}^N$ to map time-dependent inputs $\textbf{\textit{x}}(t) \in \mathbb{R}$ to outputs $\textbf{\textit{y}}(t) \in \mathbb{R}$ as follows:
\begin{equation}
\textbf{\textit{h}}'(t) = \textbf{\textit{A}} \textbf{\textit{h}}(t) +\textbf{\textit{B}} \textbf{\textit{x}}(t), \label{eq:1}
\end{equation}
\begin{equation}
\textbf{\textit{y}}_t = \textbf{\textit{C}} \textbf{\textit{h}}(t) + \textbf{\textit{D}} \textbf{\textit{x}}(t), \label{eq:2}
\end{equation}\par
\noindent \textcolor{black}{specifically, Equation (\ref{eq:1}) is the state equation, where $\textbf{\textit{A}}$ contains historical information and updates the hidden state, and $\textbf{\textit{B}}$ regulates the input's influence. Equation (\ref{eq:2}) is the output equation, where $\textbf{\textit{C}}$ projects the hidden state to the observed output $\textbf{\textit{y}}(t)$. The parameter $\textbf{\textit{D}}$ provides a direct path from input to output and can be omitted by setting $\textbf{\textit{D}} = \textbf{0}$.}\par
\textcolor{black}{Mamba is a discrete version of this system, using a time scale $\mathbf{\Delta}$ to discretize $\textbf{\textit{A}}$ and $\textbf{\textit{B}}$ into $\overline{\textbf{\textit{A}}}$ and $\overline{\textbf{\textit{B}}}$ via Zero-order Hold (ZOH):}
\begin{equation}
    \textbf{\textit{h}}_t = \overline{\textbf{\textit{A}}} \textbf{\textit{h}}_{t-1} + \overline{\textbf{\textit{B}}} \textbf{\textit{x}}_t, \quad  
 \textbf{\textit{y}}_t = \textbf{\textit{C}} \textbf{\textit{h}}_t,
\end{equation}
% \begin{equation}
%     y_t = \textbf{\textit{C}} h_t
% \end{equation}
\par
\textcolor{black}{The model’s output is computed via convolution, with M as the input sequence length $\textbf{\textit{x}}$ and $\overline{\textbf{\textit{K}}}$ as the structured convolution kernel:}
\begin{equation}
\overline{\textbf{\textit{K}}}=\left(\textbf{\textit{C}} \overline{\textbf{\textit{B}}}, \textbf{\textit{C}} \overline{\textbf{\textit{A}\textit{B}}}, \ldots, \textbf{\textit{C}} \overline{\textbf{\textit{A}}}^{\mathrm{M}-1} \overline{\textbf{\textit{B}}}\right),\quad \textbf{\textit{y}}=\textbf{\textit{x}} \ast \overline{\textbf{\textit{K}}},
\end{equation}
where $\ast$ denotes the convolution operation.
\par
\textcolor{black}{Optimized for parallel computation, models like Mamba advance SSMs in complex dynamics. However, in motion generation tasks, low-frequency components capture static postures, while high-frequency components reflect fine-grained motion. The state transition matrix $\textbf{\textit{A}}$ in Mamba defines system dynamics but lacks control over frequency-specific details crucial for enhancing motion generation.}\par
{\textbf{Latent Motion Diffusion Model}. Diffusion probabilistic models enhance motion generation by denoising a Gaussian distribution to approximate the target distribution $p(x)$ via a T-step Markov process $\{\textbf{\textit{x}}_t\}_{t=1}^T$ \cite{tevet2023human,ho2020denoising}. Our model employs a denoiser $\epsilon_\theta(\textbf{\textit{x}}_t, t)$ to iteratively reduce noise, yielding motion sequences $\left\{\hat{\textbf{\textit{x}}}_t^{1: N}\right\}_{t=1}^T$. To improve efficiency, we conduct diffusion in the latent space \cite{zhang2025motion}. Given a condition $c$, such as descriptive text $\boldsymbol{w}^{1:N}$ encoded by the frozen CLIP model $\tau_\theta^w$, we obtain the text embedding $\textbf{\textit{f}}^t=\tau_\theta^w\left(w^{1: N}\right)$, conditioning the denoiser as $\epsilon_\theta(\textbf{\textit{z}}_t, t, \tau_\theta(c))$. Using the VAE ($\mathcal{V}=\{\mathcal{E}, \mathcal{D} \}$) from MLD \cite{chen2023executing}, we project motion sequences to the latent space $\textbf{\textit{z}}=\mathcal{E}\left(\textbf{\textit{x}}^{1: L}\right)$, then decode them back as $\hat{\textbf{\textit{x}}}^{1: L}=\mathcal{D}(\textbf{\textit{z}})$. Finally, we train the model by minimizing the Mean Squared Error (MSE) between true and predicted noise in the latent space, facilitating efficient and high-quality motion generation.
\subsection{FTMoMamba Framework}
\label{sec3.2}
The overview of FTMoMamba is described in Figure \ref{fig:2}. To achieve fine-grained characterization of motion generation and ensure text-motion consistency, we propose an FTMamba consisting of a Frequency Mamba (FreqMamba) and a Text Mamba (TextMamba). FreqMamba focuses on generating static poses and fine-grained motions by utilizing low- and high-frequency information, while TextMamba ensures text-motion consistency by aligning text features. \par
% \par
In motion generation tasks, diffusion models in latent space improve the quality of motion generation by learning latent features, thereby reducing the redundancy introduced by data compression \cite{chen2023executing, huang2024stablemofusion,zhang2025motion}. However, due to the lack of frequency-domain information, the model's ability to capture static poses (\textit{e.g.}, sit, lay) and fine-grained motions (\textit{e.g.}, transition, stumble) is weak, leading to difficulties in capturing fine-grained motions.\par
\textbf{FreqMamba.} To address the problem, we propose a Frequency-domain Mamba (FreqMamba). 
% This method decomposes frequency-domain information into low- and high-frequency components, which guide the generation of static poses and fine-grained motions, respectively. 
First, we define the feature $\textbf{\textit{f}}_m^{\text{mo}}$ as the feature extracted by the convolution operation \cite{huang2024stablemofusion} in FTMamba from the noisy feature $\textbf{\textit{z}}_t$ and the time step \textit{t}. Second, FreqMamba (as shown in Figure \ref{fig:2}) projects $\textbf{\textit{f}}_m^{\text{mo}}$ to $\textbf{\textit{f}}_m^{{\text{Linear}}}$ through a linear layer. Then, it extracts the latent temporal features of $\textbf{\textit{f}}_m^{\text{Linear}}$ using cascaded depthwise convolution (CDWConv) \cite{li2023tfformer}, and employs the frequency state space model (FreqSSM) to capture high and low-frequency information, guiding the model to generate fine-grained motions. The resulting feature is defined as $\textbf{\textit{f}}_m^{\text{freq}}$. The formula is as follows:
\begin{equation}
\textbf{\textit{f}}_m^{\text{Linear}} = \texttt{Linear}(\textbf{\textit{f}}_m^{\text{mo}}),
\end{equation}
\begin{equation}
\textbf{\textit{f}}_m^{\text{freq}} = \texttt{FS}(\sigma(\texttt{CDWConv}(\textbf{\textit{f}}_m^{\text{Linear}}))) ,
\end{equation}
where $\sigma$ is the sigmoid activation function. $\texttt{Linear}$ ($\cdot$) is a linear layer. $\texttt{CDWConv}$($\cdot$) is the cascaded depthwise convolution, which uses a kernel size of 3 and an increasing dilation factor of (1, 2, 4) to enlarge the convolutional receptive field and capture shallow temporal features. $\texttt{FS}$ ($\cdot$) is the FreqSSM. We further perform  information selection through a multiplication-based method to obtain the motion feature $ \textbf{\textit{f}}_n^{\text{mo}}$ guided by frequency-domain information:
\begin{equation}
\textbf{\textit{f}}_n^{\text{mo}} = \texttt{Linear}(\textbf{\textit{f}}_m^{\text{freq}} \odot \sigma(\textbf{\textit{f}}_m^{\text{Linear}})) ,
\end{equation}
where $\odot$ is the Hadamard product.
\par
\begin{figure}[!t]
    \centering
    \includegraphics[width=1\linewidth]{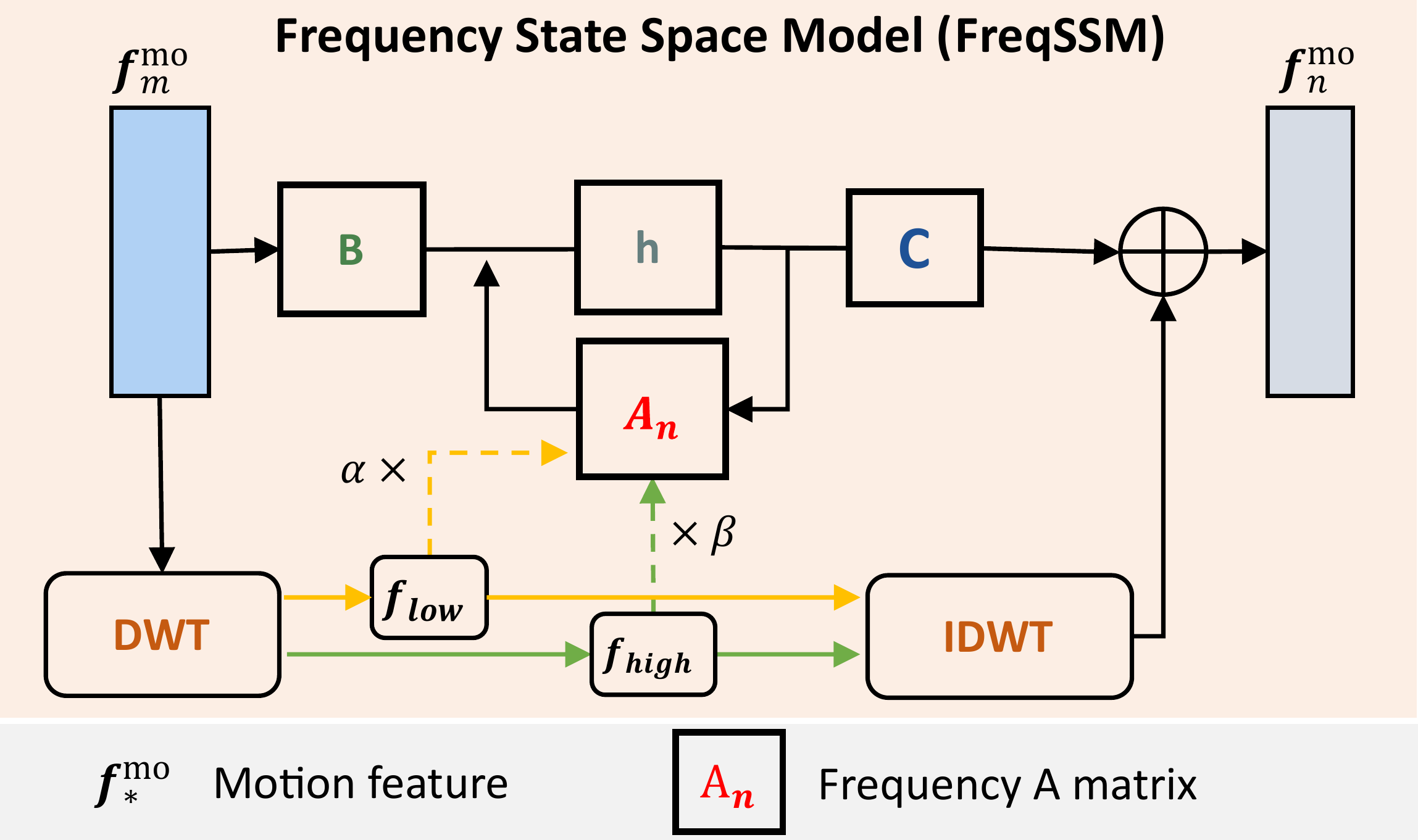}
    \caption{Architecture of FreqSSM. It decomposes motion features into low-frequency ($\textbf{\textit{f}}_{\text{low}}$) and high-frequency ($\textbf{\textit{f}}_{\text{high}}$) features. Then, it reconstructs the state transition matrix $\textbf{\textit{A}}_n$ using the frequency-domain features.}
    \label{fig:3}
    \vspace{-3mm}
\end{figure}

\textbf{FreqSSM.} Inspired by \cite{chang2010frequency} that uses high-frequency information to recognize dynamic motions and \cite{kothandaraman2022differentiable} that leverages low-frequency information to identify non-significant regions, we propose a Frequency State Space Model (FreqSSM). As shown in Figure \ref{fig:3}, FreqSSM captures static poses with low-frequency information and guides fine-grained with high-frequency information to achieve fine-grained generation. First, we decompose sequence features $\textbf{\textit{f}}_m^{\text{mo}}$  into low-frequency and high-frequency components using Discrete Wavelet Transform (DWT). After convolutional feature enhancement, we obtain \( \textbf{\textit{f}}_{\text{low}} \) and \( \textbf{\textit{f}}_{\text{high}} \). Then, based on the original state transition matrix \( \textbf{\textit{A}} \), we dynamically adjust the proportion of low- and high-frequency information in the matrix using learnable parameters. This results in the frequency-domain state transition matrix \( \textbf{\textit{A}}_n \), which then guides state updates. The $\textbf{\textit{A}}_n$ can be written as:
\begin{equation}
(\textbf{\textit{f}}_{\text{low}}, \textbf{\textit{f}}_{\text{high}}) = \texttt{Conv}(\texttt{DWT}(\textbf{\textit{f}}_m^{\text{mo}})) ,
\end{equation}
\begin{equation}
\textbf{\textit{A}}_n = \textbf{\textit{A}} + \alpha \textbf{\textit{f}}_{\text{low}} + \beta \textbf{\textit{f}}_{\text{high}} ,
\end{equation}
where $\texttt{DWT}$($\cdot$) is the discrete wavelet transform. $\texttt{Conv}$($\cdot$) is the convolution, which uses a kernel size of 3. $\alpha$ and $\beta$ are learnable parameters, which are automatically updated during training through backpropagation to achieve a dynamic balance between low- and high-frequency information. \par
Then, via $ \textbf{\textit{A}}_n$ as the core, we combine it with matrix $\textbf{\textit{B}}$ to update the hidden state matrix \( \textbf{\textit{h}} \), and obtain the state space model output through the observation matrix. Finally, \( \textbf{\textit{f}}_{\text{low}} \) and \( \textbf{\textit{f}}_{\text{high}} \) are combined with the state space model output in a residual manner. After being transformed back via Inverse Discrete Wavelet Transform (IDWT), they incorporate frequency-domain enhanced information to achieve fine-grained motion generation:
\begin{equation}
\textbf{\textit{h}}_t' = \textbf{\textit{A}}_n \textbf{\textit{h}}_t + \textbf{\textit{B}} \textbf{\textit{f}}_m^{\text{mo}} ,
\end{equation}
\begin{equation}
\textbf{\textit{f}}_n^{\text{mo}} = \textbf{\textit{C}} \textbf{\textit{h}}_t + \texttt{IDWT}(\textbf{\textit{f}}_{\text{low}}, \textbf{\textit{f}}_{\text{high}}) ,
\end{equation}
where $\texttt{IDWT}$($\cdot$) is the inverse discrete wavelet transform.
\par
In text-driven human motion generation tasks, the generated motion is directly constrained by the text instructions. Therefore, effectively understanding the semantic information of the text and achieving precise alignment between semantics and motion is crucial for generating motions \cite{zhang2025motion,chen2023executing}. However, the above method directly concatenates text features and motion features. Such simple concatenation introduces text-motion semantic inconsistency as the motions become more complex.

% Therefore, we aim to optimize the design of text-conditioned guidance to accurately align the texts and motions during the denoising process of the Diffusion model.
\par
\begin{figure}[!t]
    \centering
    \includegraphics[width=1\linewidth]{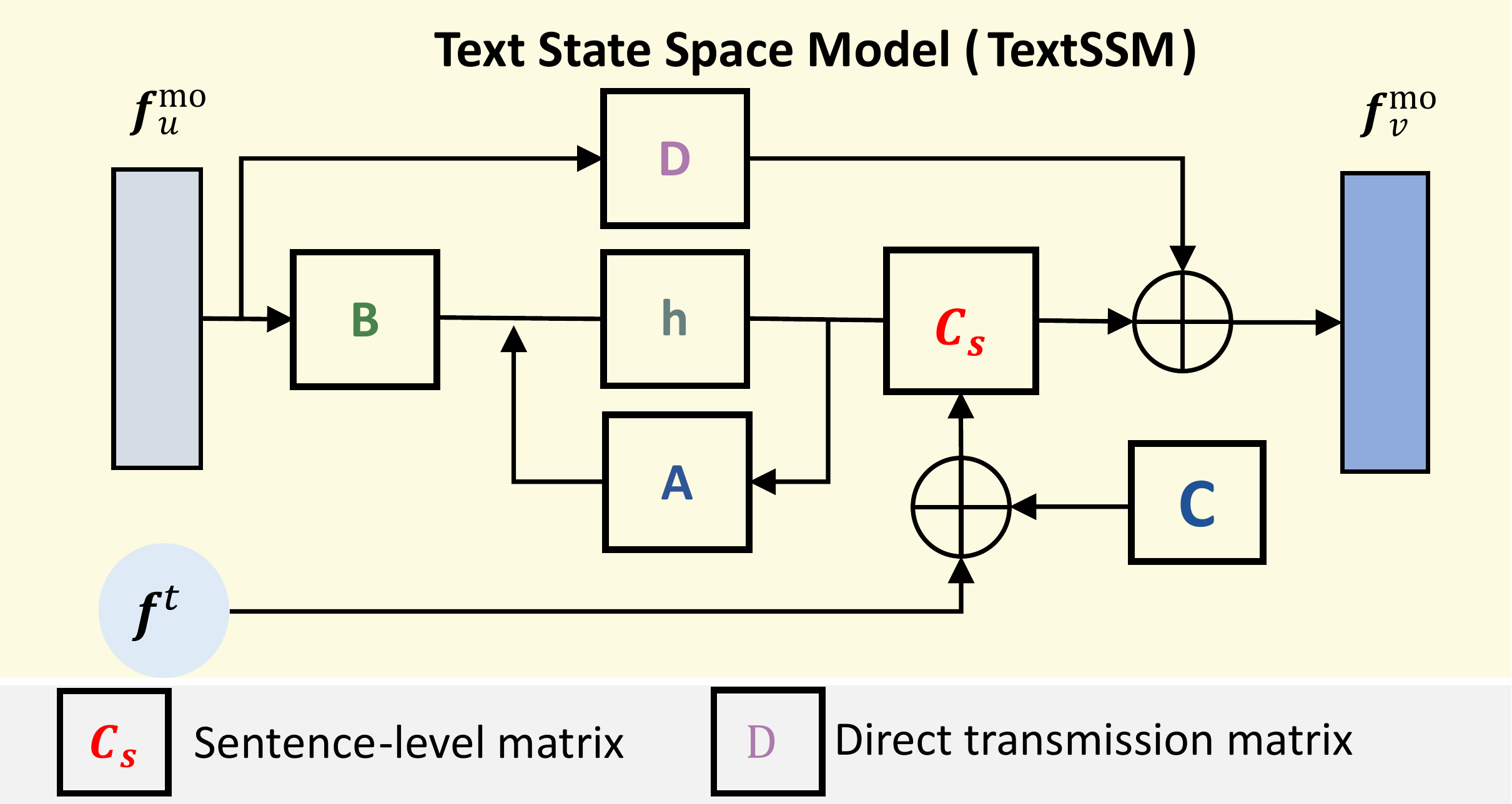}
    \caption{Architecture of TextSSM. TextSSM integrates the text features $\textbf{\textit{f}}^t$ and the output matrix \textbf{\textit{C}}. Under negligible computational cost, we reconstruct the Text State Space Model to enable cross-modal fusion.}
    \label{fig:4}
\end{figure}

\textbf{TextMamba. }To optimize the design of text-conditioned guidance to accurately align the texts and motions, we propose a TextMamba. First, we define the feature $\textbf{\textit{f}}_u^{\text{mo}}$ as the residual connection of $\textbf{\textit{z}}_t$ and $\textbf{\textit{f}}_n^{\text{mo}}$. Second, TextMamba (as shown in Figure \ref{fig:2}) projects $\textbf{\textit{f}}_u^{\text{mo}}$ to $\textbf{\textit{f}}_u^{{\text{Linear}}}$ by a linear layer. Then, it extracts the latent temporal features of $\textbf{\textit{f}}_u^{\text{Linear}}$ using CDWConv, and  TextSSM aligns the text and motion features. The resulting feature is defined as $\textbf{\textit{f}}_u^{\text{text}}$ as follows:
\begin{equation}
\textbf{\textit{f}}_u^{\text{Linear}} = \texttt{Linear}(\textbf{\textit{f}}_u^{\text{mo}}),
\end{equation}
\begin{equation}
\textbf{\textit{f}}_u^{\text{text}} = \texttt{TS}(\sigma(\texttt{CDWConv}(\textbf{\textit{f}}_u^{\text{Linear}}))) ,
\end{equation}
where $\texttt{TS}$($\cdot$) is the TextSSM. Furthermore, we apply a multiplication-based method for information selection to obtain the text-motion alignment feature $\textbf{\textit{f}}_v^{\text{mo}}$, guided by sentence-level text information:
\begin{equation}
\textbf{\textit{f}}_v^{\text{mo}} = \texttt{Linear}(\textbf{\textit{f}}_u^{\text{text}} \odot \sigma(\textbf{\textit{f}}_u^{\text{Linear}})) ,
\end{equation}
\par
\begin{table*}[!t]
% \belowrulesep=0pt
% \aboverulesep=0pt
\centering

  \setlength{\tabcolsep}{0.17cm}

\begin{tabular}{cccccccc}
\toprule
                          & \multirow{2}{*}{FID $\downarrow$}       & \multicolumn{3}{c}{R-Precision $\uparrow$}                                                             &                             &                               &                               \\ \cline{3-5}
\multirow{-2}{*}{Methods} &                             & Top-1                         & Top-2                         & Top-3                         & \multirow{-2}{*}{MM-Dist $\downarrow$} & \multirow{-2}{*}{Diversity $\uparrow$} & \multirow{-2}{*}{MModality $\uparrow$} \\ \midrule
Real motions              
& $0.002^{\pm.000}$            & $0.511^{\pm .003}$                   & $0.703^{\pm.003}$                    & $0.797^{\pm .002}$                  & $2.974^{\pm .008}$                           & $9.503^{\pm.065}$                    & -                             \\ \hline
Hier \cite{ghosh2021synthesis} &
$6.532^{\pm.024}$             & $0.301^{\pm.002}$ & $0.425^{\pm.002}$ & $0.552^{\pm.004}$ & $5.012^{\pm.018}$ & $8.332^{\pm.042}$ & \multicolumn{1}{c}{-} \\
TEMOS \cite{petrovich2022temos} &
$3.734^{\pm.028}$             & $0.424^{\pm.002}$ & $0.612^{\pm.002}$ & $0.722^{\pm.002}$ & $3.703^{\pm.008}$ & $8.973^{\pm.071}$ & $0.368^{\pm.018}$ \\
TM2T\cite{guo2022tm2t} 
&$1.501^{\pm.017}$             & $0.424^{\pm.003}$ & $0.618^{\pm.003}$ & $0.729^{\pm.002}$ & $3.467^{\pm.011}$ & $8.589^{\pm.076}$ & $2.424^{\pm.093}$ \\
T2M \cite{guo2022generating} &
$1.067^{\pm.002}$             & $0.457^{\pm.002}$ & $0.639^{\pm.003}$ & $0.740^{\pm.003}$ & $3.340^{\pm.008}$ & $9.188^{\pm.002}$ & $2.090^{\pm.083}$ \\
MDM \cite{tevet2023human} &
${0.544}^{\pm.044}$            & $0.320^{\pm.005}$ & $0.498^{\pm.004}$ & $0.611^{\pm.007}$ & $5.566^{\pm.027}$ & $9.559^{\pm.086}$ & $\mathbf{2.799}^{\pm.072}$ \\
MotionDiffuse \cite{zhang2024motiondiffuse} &
$0.630^{\pm.001}$             & ${0.491}^{\pm.001}$ & ${0.681}^{\pm.001}$ & ${0.782}^{\pm.001}$ & ${3.113}^{\pm.001}$ & ${9.410}^{\pm.049}$ & $1.553^{\pm.042}$ \\
$\text{MLD}^*$ \cite{chen2023executing} &
\color{black}{$0.421^{\pm.010}$} & \color{black}{$0.468^{\pm.002}$ }                        & \color{black}{$0.658^{\pm.002}$}                         & \color{black}{$0.759^{\pm.002}$}                         &  \color{black}{$3.267^{\pm.009}$}                          & \color{black}{$9.701^{\pm.072}$}                         & \color{black}{$2.600^{\pm.106}$}                         \\
Motion Mamba \cite{zhang2025motion}   &
${0.281}^{\pm.009}$             & $\mathbf{0.502}^{\pm.003}$ & $\mathbf{0.693}^{\pm.002}$ & $\mathbf{0.792}^{\pm.002}$ & $\mathbf{3.060}^{\pm.058}$ & $\mathbf{9.871}^{\pm.084}$ & $\underline{2.294}^{\pm.058}$                          \\

Fg-T2M \cite{wang2023fg} 
&$0.243^{\pm.019}$             & $\underline{0.492}^{\pm.002}$ & {$\underline{0.683}^{\pm.003}$} & {$\underline{0.783}^{\pm.002}$ } & $3.109^{\pm.007}$ & $9.278^{\pm.072}$  & $1.614^{\pm.049}$ \\
MotionGPT \cite{jiang2023motiongpt}           
& $\underline{0.232}^{\pm.008}$ & $0.492^{\pm.003}$                         & $0.681^{\pm.003}$                         & $0.778^{\pm.002}$                         & {$\underline{3.096}^{\pm.008}$}                            & \color{black}{$9.528^{\pm.071}$ }                          & $2.008^{\pm.084}$       \\ 
\hline
FTMoMamba(Ours)                      
& { $\mathbf{0.181}^{\pm.009}$} 
& {\color{black} $0.489^{\pm.003}$} 
& {\color{black} $0.680^{\pm.002}$} 
& {\color{black} $0.777^{\pm.002}$} 
&        $3.151^{\pm.009}$                     
& { $\underline{9.789}^{\pm.085}$}   
& {\color{black} $2.277^{\pm.099}$}  \\ \bottomrule

\end{tabular}
\caption{Quantitative Comparison on HumanML3D test set. For each metric, we repeat the evaluation 20 times and report the average with 95\% confidence interval. Boldface indicates the best result, while \underline{underscore} refers to the suboptimal result. * denotes the results obtained using the official weight testing.}
  \label{table1}
\end{table*}
\textbf{TextSSM.} Inspired by the cross-attention information fusion method, we design a Text State Space Model (TextSSM). As shown in Figure \ref{fig:4}, TextSSM combines sentence-level features with the output matrix \( \textbf{\textit{C}} \) of the state space model, achieving cross-modal semantic alignment and ensuring text-motion consistency. In TextSSM, we first compute the state equation for the motion feature \( \textbf{\textit{f}}_u^{\text{mo}}\), updating the hidden state. Then, the sentence-level feature \( \textbf{\textit{f}}^{\text{t}} \) extracted by CLIP is summed with the output matrix \( \textbf{\textit{C}} \), achieving text-to-motion alignment with minimal computational cost. The formula is as follows:
\begin{equation}
\textbf{\textit{C}}_s = \textbf{\textit{f}}^\text{t} + \textbf{\textit{C}} ,
\end{equation}
where $\textbf{\textit{C}}_s$ is the sentence-level output martix. Finally, the aligned features are obtained through the observation matrix, ensuring consistency between text and motion. The formulas can be written as :
\begin{equation}
\textbf{\textit{h}}_t' = \textbf{\textit{A}} \textbf{\textit{h}}_t + \textbf{\textit{B}} \textbf{\textit{f}}_u^{\text{mo}} ,
\end{equation}
\begin{equation}
\textbf{\textit{f}}_v^{\text{mo}} = \textbf{\textit{C}}_s \textbf{\textit{h}}_t + \textbf{\textit{D}} \textbf{\textit{f}}_u^{\text{mo}},
\end{equation}\par
\noindent Finally, the proposed FTMamba can be denoted as:
\begin{equation}
{\textbf{\textit{z}}_t^{'}}=\textbf{\textit{f}}_n^{\text{mo}}+\textbf{\textit{f}}_v^{\text{mo}} ,
\end{equation}\par
\noindent where $\textbf{\textit{z}}_t^{'}$ denotes the encoded feature of FTMamba. 
\section{Expereiments}

\subsection{Dataset}
Following \cite{chen2023executing,zhang2024motiondiffuse,zhang2025motion,wang2023fg}, we evaluate our proposed FTMoMamba on one widely used motion-language benchmark, HumanML3D \cite{guo2022generating}. The HumanML3D dataset collects 14616 motions from AMASS \cite{mahmood2019amass} and HumanAct12 \cite{guo2020action2motion} datasets, with each motion described by 3 textual scripts, totaling 44970 descriptions. It also contains a variety of actions, including exercising, dancing, and acrobatics. 
\subsection{Evaluation Metrics}
We adopt standard evaluation metrics across the following aspects in our experiments: (1) \textit{Fréchet Inception Distance} (\textbf{FID}), which measures the overall quality of the generated motions by quantifying the distributional difference between the high-level features of the generated motions and those of real motions; (2) \textit{R-Precision} and \textit{Multimodal Distance} (\textbf{MM-Dist}), which evaluate the semantic alignment between the input text and the generated motions; (3) \textbf{Diversity}, which measures motion diversity; and (4) \textit{Multimodality} (\textbf{MModality}), which assesses the diversity of motions generated from the same input text.
\subsection{Implementation Details}
For the comparisons, the encoder and decoder use the same configuration and weights as in MLD \cite{chen2023executing}. The text embedding $\textbf{\textit{f}}^t \in \mathbb{R}^{1 \times 256}$ and the latent $\textbf{\textit{z}} \in \mathbb{R}^{16 \times 256}$ for diffusion learning and inference. In the denoising UNet architecture, we set the FTMamba to have $2$ layers with $256$ channels in both the encoder, middle, and decoder stages. We use a {CLIP-ViT-L-$14$} \cite{radford2021learning} model with frozen weights as the text encoder for text condition. In the experiment, we use the bidirectional model of FreqMamba to extract and utilize frequency-domain information. The information fusion is performed through forward and backward summation. In the ablation experiments, the same number of model layers, inputs, and environmental configurations as in FTMoMamba were used. All our models are trained with the AdamW optimizer using a fixed learning rate of $10^{{-}4}$. Our batch size is set to  $64$ during the diffusion stage, and the model is trained for $2000$ epoch. The number of diffusion steps is $1000$ during training while $50$ during interfering, and the variances $\beta_t$ are scaled linearly from $ 8.5 \times 10^{{-}4}$ to $0.012$. The model is trained on two RTX 3090 GPUs and tested on a single RTX 3090 GPU.

\begin{figure*}[!t]
    \centering
    \includegraphics[width=1\linewidth]{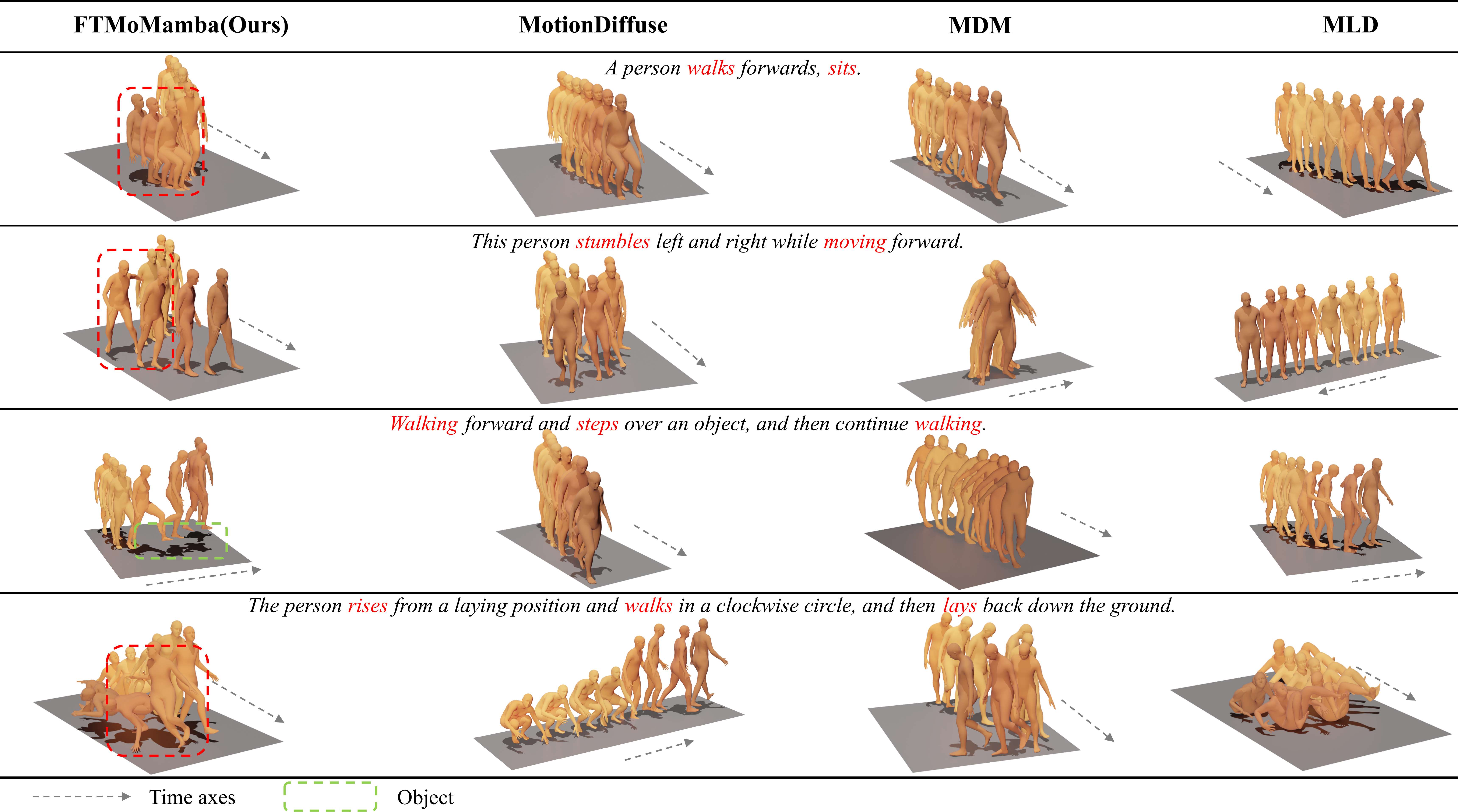}
    \caption{Qualitative comparison on HumanML3D test dataset.}
    \vspace{-2mm}
    \label{fig:5}
\end{figure*}

\begin{table*}[!t]
\centering
 
  \setlength{\tabcolsep}{0.12cm}

\begin{tabular}{ccccccccc}
\toprule
                      &                         &                                                             &       \multicolumn{3}{c}{R-Precision $\uparrow$}                         &                             &                               &                               \\ \cline{4-6}
\multirow{-2}{*}{FreqMamba} & \multirow{-2}{*}{TextMamba} & \multirow{-2}{*}{FID $\downarrow$}   & \multicolumn{1}{c}{\centering Top-1} & \multicolumn{1}{c}{\centering Top-2} & \multicolumn{1}{c}{\centering Top-3} & \multirow{-2}{*}{MM-Dist $\downarrow$} & \multirow{-2}{*}{Diversity $\uparrow$} & \multirow{-2}{*}{MModality $\uparrow$} \\ \midrule
$\usym{2718}$                     & $\usym{2718}$    
& $0.263^{\pm.009}$                         
& $0.489^{\pm.002}$ 
& $0.676^{\pm.002}$ 
& $0.772^{\pm.003}$
& $3.162^{\pm.008}$ 
& $9.679^{\pm.066} $
& $2.268^{\pm.091}$ \\

$\usym{2718}$                      &   $\usym{2713}$  
& $\underline{0.209}^{\pm.006}$                      
& $0.483^{\pm.003} $
& $0.672^{\pm.003}$ 
& $0.768^{\pm.002}$ 
& $3.198^{\pm.009}$ 
& $\mathbf{9.841}^{\pm.092}$ 
& $\underline{2.288}^{\pm.088}$     \\

$\usym{2713}$                      & $\usym{2718}$    
& $0.227^{\pm.009}$                        
& $\mathbf{0.497}^{\pm.003} $
& $\mathbf{0.687}^{\pm.003}$ 
& $\mathbf{0.782}^{\pm.002}$
& $\mathbf{3.117}^{\pm.006}$
& $9.775^{\pm.086}$
& $\mathbf{2.29}^{\pm.093}$ \\

$\usym{2713}$                      & $\usym{2713}$    & { $\mathbf{0.181}^{\pm.009}$} 
& {$\underline {0.489}^{\pm.003}$} 
& $\underline {0.680}^{\pm.002}$
& {$\underline {0.777}^{\pm.002}$} 
&  $\underline{3.151}^{\pm.009}$                    
& {$\underline{9.789}^{\pm.085}$}
& {\color{black} $2.277^{\pm.099}$} \\ \bottomrule
\end{tabular}
 \caption{Ablation experiment of FreqMamba and TextMamba modules on the HumanML3D test dataset.}
 \vspace{-2mm}
   \label{table2}
\end{table*}

\subsection{Quantitative and Qualitative Comparison}
\noindent{\bf Quantitative Comparison.} As shown in Table \ref{table1}, we observe that our method achieves significant improvements over the baseline method (\textit{i.e.}, MLD \cite{chen2023executing}) in terms of R-Precision, FID, MM-Dist, Diversity, and MModality. Compared to the state-of-the-art method MotionGPT \cite{jiang2023motiongpt}, the performance gap in other metrics is negligible, while the FID score is reduced by 5.1\%, which demonstrates the advantage of the proposed method in terms of motion generation quality. Based on these results, we can draw the following conclusions: 1) the introduction of frequency-domain information indeed brings gains, helping the model to better capture fine-grained human motion. 2) the improvement in R-Precision further validates the effectiveness of TextSSM.\par
\noindent{\bf Qualitative Comparison.} Figure \ref{fig:5} presents qualitative comparisons of our method with MotionDiffuse \cite{zhang2024motiondiffuse}, MDM \cite{tevet2023human}, and MLD \cite{chen2023executing}. From Figure \ref{fig:5}, we can observe that MDM \cite{tevet2023human} is capable of generating simple semantic motions but struggles with capturing fine-grained transitions between sequential motions. Despite MLD \cite{chen2023executing} and MotionDiffuse \cite{zhang2024motiondiffuse} show improvements in this motion, they still fall short in accurately aligning with textual descriptions. For the instruction \textit{``This person stumbles left and right while moving forward."}, MLD \cite{chen2023executing} fails to maintain the intended forward movement. For the instruction \textit{``Walking forward and steps over an object, and then continue walking."}, both MLD \cite{chen2023executing} and MotionDiffuse \cite{zhang2024motiondiffuse} don't generate motion related to stepping over the object. In the sequence of motion \textit{``rise-walk-lay"}, each of the three methods omits at least one essential motion in the transition. From these observations, we can conclude that: 1) Frequency information introduced by FreqSSM can assist the model in fine-grained control, such as sitting, forward movement, and rotation. 2) TextSSM ensures semantic alignment, enabling the model to generate and avoid obstacles accurately. 

\begin{table*}[!t]
\centering
\setlength{\tabcolsep}{0.275cm}
\begin{tabular}{ccccccccc}
\toprule
\multirow{2}{*}{Methods} & \multirow{2}{*}{FID $\downarrow$} & \multicolumn{3}{c}{R-Precision $\uparrow$} & \multirow{2}{*}{MM-Dist $\downarrow$} & \multirow{2}{*}{Diversity $\uparrow$} & \multirow{2}{*}{MModality $\uparrow$} \\ \cline{3-5}
                                 &                                  & \centering Top-1 & \centering Top-2 & \centering Top-3 &                               &                               &                               \\ \midrule
FFT &  ${0.291}^{\pm.009}$ & ${0.486}^{\pm.002}$  & ${0.676}^{\pm.003}$ & ${0.772}^{\pm.003}$ & ${3.163}^{\pm.011}$ & ${9.719}^{\pm.073}$ &$\mathbf{2.325}^{\pm.100}$
\\
DWT                    & {$\mathbf {0.181}^{\pm.009}$}     & {$\mathbf{0.489}^{\pm.003}$} & $\mathbf{ 0.680}^{\pm.002}$ & {$\mathbf{0.777}^{\pm.002}$} & {$\mathbf{3.151}^{\pm.009}$} & { $\mathbf{9.789}^{\pm.085}$} & { $\underline{2.277}^{\pm.099}$} \\ \bottomrule
\end{tabular}
\caption{Ablation experiment of different frequency-domain information extraction methods on the HumanML3D test dataset.}
\label{table4}
\end{table*}

\begin{table*}[t]
\centering

  \setlength{\tabcolsep}{0.25cm}

\begin{tabular}{ccccccccc}
\toprule
                      &                         &                                                            &  \multicolumn{3}{c}{R-Precision $\uparrow$}                               &                             &                               &                               \\ \cline{4-6}
\multirow{-2}{*}{$\textbf{\textit{f}}_{\text{low}}$} & \multirow{-2}{*}{$\textbf{\textit{f}}_{\text{high}}$} & \multirow{-2}{*}{FID $\downarrow$}   & \multicolumn{1}{c}{\centering Top-1} & \multicolumn{1}{c}{\centering Top-2} & \multicolumn{1}{c}{\centering Top-3} & \multirow{-2}{*}{MM-Dist $\downarrow$} & \multirow{-2}{*}{Diversity $\uparrow$} & \multirow{-2}{*}{MModality $\uparrow$} \\ \midrule
$\usym{2718}$                     & $\usym{2718}$    
& {$\underline{0.209}^{\pm.006}$}                        
& $0.483^{\pm.003} $
& $0.672^{\pm.003}$ 
& $0.768^{\pm.002}$ 
& $3.198^{\pm.009}$ 
& $\underline{9.841}^{\pm.092}$ 
& $2.288^{\pm.088}$ \\

$\usym{2718}$                      &   $\usym{2713}$ 
& $0.232^{\pm.005}$                       
& $0.489^{\pm.003} $
& $\mathbf {0.682}^{\pm.002}$ 
& $\mathbf {0.780}^{\pm.002}$
& $\mathbf {3.150}^{\pm.013}$ 
& $\mathbf {9.945}^{\pm.066}$
& {$\underline{2.325}^{\pm.081}$}                        \\

$\usym{2713}$                      & $\usym{2718}$    
& $0.216^{\pm.006}$                         
& $\mathbf{0.490}^{\pm.002}$
& $0.682^{\pm.003} $
& {$\underline{0.779}^{\pm.002}$ }
& $3.160^{\pm.010} $
& $9.818^{\pm.085} $
& $\mathbf {2.399}^{\pm.088}$     \\

$\usym{2713}$                      & $\usym{2713}$   
& {$\mathbf {0.181}^{\pm.009}$} 
& {$\underline{0.489}^{\pm.003}$}
& $\underline{ 0.680}^{\pm.002}$
& {\color{black} $0.777^{\pm.002}$} 
& {$\underline{3.151}^{\pm.009}$}                     
& {\color{black} $9.789^{\pm.085}$}   
& {\color{black} $2.277^{\pm.099}$}  \\ \bottomrule
\end{tabular}
  \caption{Ablation experiment of low-frequency ($\textbf{\textit{f}}_{\text{low}}$) and high-frequency ($\textbf{\textit{f}}_{\text{high}}$) information on the HumanML3D test dataset.}
    \label{table3}
\end{table*}

\begin{figure*}[!t]
    \centering
    \includegraphics[width=1\linewidth]{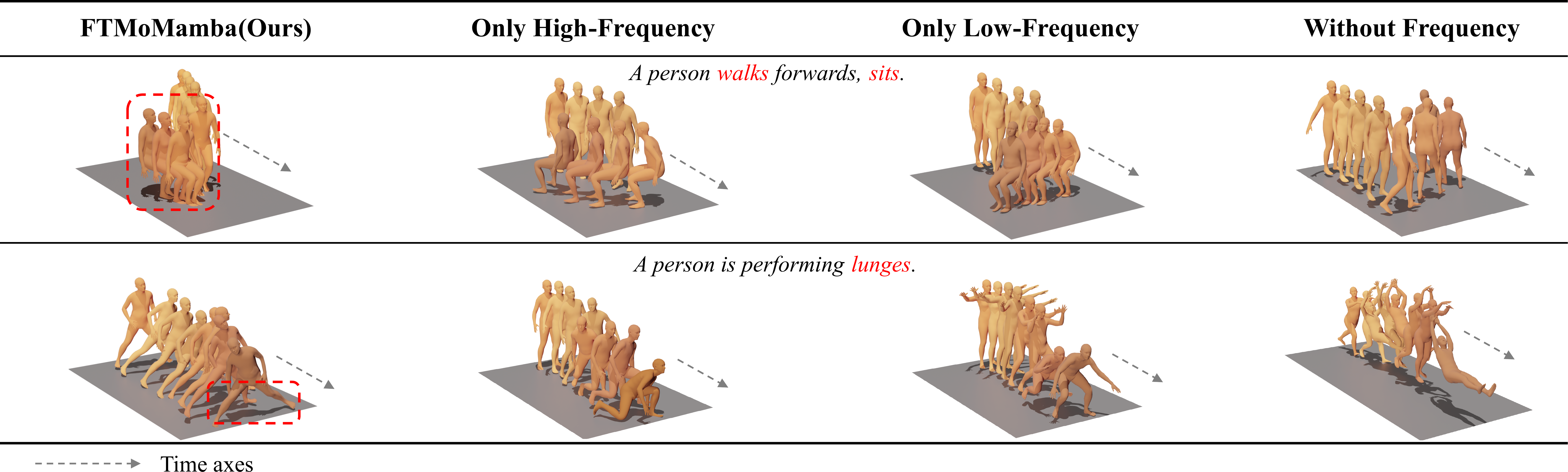}
    \caption{Visualization of our method with only frequency-domain information on the HumanML3D test dataset.}
    \label{fig:6}
    % \vspace{-0cm}
\end{figure*}

\subsection{Ablation Studies}
\noindent{\bf Effectiveness of FreqMamba and TextMamba.} When FreqMamba is $\protect\usym{2718}$, use BiMamba instead; When TextMamba is $\protect\usym{2718}$, apply Linear Attention instead. From Table \ref{table2}, when either FreqMamba or TextMamba is used alone, significant improvements in FID, Diversity, and MModality are observed compared to the baseline, indicating that frequency domain information or sentence-level textual features can effectively enhance the quality of motion generation. A comparison of the experimental results reveals that TextMamba outperforms FreqMamba in terms of lower FID and greater diversity. The primary reason is that text alignment enhances the model's understanding of the text, improving the generation quality. When both FreqMamba and TextMamba are used together, the lowest FID of 0.181 is achieved, demonstrating the complementarity of frequency domain guidance and text-motion alignment.\par

\noindent{\bf Effectiveness of different frequency-domain information extraction methods.} From Table \ref{table4}, we can observe that DWT achieves significant improvements across multiple metrics. The main reason is that the Fast Fourier Transform (FFT) only extracts frequency domain information without directly distinguishing between low and high frequencies, leading to information confusion. In this case, the frequency domain information guiding static poses and fine-grained motion generation is severely disrupted. Moreover, the DWT’s strategy of separating low- and high-frequency extraction, along with the dynamic balancing of both components using learnable parameters, effectively prevents this issue.\par

\noindent{\bf Impact of low-frequency and high-frequency information.} From Table \ref{table3} and Figure \ref{fig:6}, we observe that using both low- and high-frequency information leads to improvements in R-Precision, MM-Dist, and MModality compared to the baseline. This indicates that both low- and high-frequency components can enhance text-motion consistency and increase the diversity of generated motions. The decrease in FID suggests that a single type of frequency domain information leads to an imbalance in the information. Finally, when both low- and high-frequency information are used together, FID decreases by 2.8\% relative to the baseline, without compromising text-motion consistency. This further demonstrates their complementarity and the reduction of redundant information interference via learnable parameters.
\section{Conclusion}
In this work, we propose a novel FTMoMamba to generate human motions based on text instructions. It has two key techniques: FreqSSM is responsible for fine-grained motion generation, and TextSSM ensures text-motion consistency. In addition, FTMoMamba is efficient and flexible, which can accurately generate high-quality motions. Extensive qualitative and quantitative experiments show that FTMoMamba outperforms SOTA methods. Although our method achieves the lowest FID (\textit{i.e.}, directly validating the quality of motion generation), the lack of foot contact loss, pose loss, and other physical-related losses results in less significant advantages in other evaluation metrics. In the future, we will consider incorporating these losses with physical prior, ensuring further improvement of other performance metrics.

{
    \small
    \bibliographystyle{ieeenat_fullname}
    \bibliography{main}
}

\end{document}